\documentclass[final,5p,times,twocolumn]{elsarticle}

\usepackage{graphics} 
\usepackage{graphicx} 
\usepackage{epsfig}
\usepackage{amssymb}
\usepackage{gensymb}
\usepackage{tabularx}
\usepackage{multirow}
\usepackage{amssymb}
\usepackage{amsmath}
\usepackage{eurosym}
\biboptions{numbers,sort&compress}
\usepackage{subfigure}
\usepackage{arydshln}
\usepackage{lineno,hyperref}
\usepackage{float}
\usepackage{threeparttable}
\usepackage{adjustbox}
\usepackage{verbatim} 
\usepackage[english]{babel}

\journal{XXXX-XXXX}

\begin{document}

\begin{frontmatter}

\title{Supervised learning for improving the accuracy of robot-mounted 3D camera applied to human gait analysis.}

\author[UPM]{Diego Guffanti}
\ead{d.guffanti@alumnos.upm.es}

\author[UPM]{Alberto Brunete}
\author[UPM]{Miguel Hernando}
\author[ETSIDI]{David \'Alvarez}
\author[INEF]{Javier Rueda}
\author[INEF]{Enrique Navarro}

\address[UPM]{Centre for Automation and Robotics (CAR UPM-CSIC), Universidad Polit\'ecnica de Madrid, 28012 Madrid, Spain.\\}
\address[ETSIDI]{Department of Electrical, Electronic and Automation Engineering and Applied Physics, ETSIDI, Universidad Polit\'ecnica de Madrid, 28012 Madrid, Spain.\\}
\address[INEF]{Department of Human Health and Performance, Faculty of Sports Sciences, Universidad Polit\'ecnica de Madrid, 28040 Madrid, Spain.\\}

\begin{abstract}
The use of 3D cameras for gait analysis has been highly questioned due to the low accuracy they have demonstrated in the past. The objective of the study presented in this paper is to improve the accuracy of the estimations made by robot-mounted 3D cameras in human gait analysis by applying a supervised learning stage. The 3D camera was mounted in a mobile robot to obtain a longer walking distance. This study shows an improvement in detection of kinematic gait signals and gait descriptors by post-processing the raw estimations of the camera using artificial neural networks trained with the data obtained from a certified Vicon system. To achieve this, 37 healthy participants were recruited and data of 207 gait sequences were collected using an Orbbec Astra 3D camera. 
There are two basic possible approaches for training and both have been studied in order to see which one achieves a better result. The artificial neural network can be trained either to obtain more accurate kinematic gait signals or to improve the gait descriptors obtained after initial processing. The former seeks to improve the waveforms of kinematic gait signals by reducing the error and increasing the correlation with respect to the Vicon system. The second is a more direct approach, focusing on training the artificial neural networks using gait descriptors directly. The accuracy of the 3D camera to objectify human gait was measured before and after training. In both training approaches, a considerable improvement was observed. Kinematic gait signals showed lower errors and higher correlations with respect to the ground truth.  The accuracy of the system to detect gait descriptors also showed a substantial improvement, mostly for kinematic descriptors rather than spatio-temporal. When comparing both training approaches, it was not possible to define which was the absolute best. Therefore, we believe that the selection of the training approach will depend on the purpose of the study to be conducted. This study reveals the great potential of 3D cameras and encourages the research community to continue exploring their use in gait analysis.

\end{abstract}

\begin{keyword}

3D camera \sep  mobile robot \sep gait analysis \sep machine learning

\end{keyword}

\end{frontmatter}

\section{Introduction}
\label{sec: Introduction}

3D cameras have been widely used for gait analysis applications. Most studies claim that these sensors are not accurate enough to be used for gait analysis \cite{Pfister:2014, Mentiplay:2015}. The low accuracy is more noticeable when analyzing gait kinematics than spatio-temporal parameters \cite{Springer:2016}. There are several factors that can influence the accuracy of these sensors: light conditions, sensor position, occlusion of different parts of the body, among others \cite{LemkensW:2013}. However, the main advantage of these sensors is undoubtedly their low cost and mode of operation. As they are markerless systems, they greatly facilitate the work of clinicians when analyzing gait and avoid disturbing patients when placing markers.

Machine Learning (ML) is an alternative that has opened the gap to improve the accuracy of 3D cameras. However, current approaches have only applied ML together with 3D cameras for the classification of normal and pathological gait \cite{SHRIVASTAVA:2021}, gait recognition \cite{HZhen:2018}, or for the classification of different pathologies \cite{Li:2019}. On the contrary, they have not pursued an enhancement of the accuracy of these sensors. This is highly relevant because for clinicians dealing with this type of gait pathologies, the classification of the disease is far less relevant than the quantifiable study of its evolution. For this reason, as a contribution to current approaches, this study aims to improve the accuracy of the estimated kinematic gait signals (joint angles) and gait descriptors when using 3D cameras. For this purpose, artificial neural networks (ANNs) are used. ANNs allow the system to learn from a gold standard system through supervised learning. It should be emphasized that in this study, ANNs do not learn a specific gait pattern. On the contrary, this study attempts to correct the measurement errors of the 3D camera based on the examples provided by the gold standard system. To achieve this, a windowing process is applied to the data and each window represents a training sample. This means that the ANN never learns the gait pattern as such. Consequently, it is possible to apply the training result to the analysis of new gait patterns obtaining better estimations.

In this study, neural networks were trained based on two criteria: training based on kinematic gait signals, and training based on gait descriptors.  The former seeks to improve the waveforms of kinematic signals in sagittal, frontal and transverse planes, by reducing the error and increasing the correlation with respect to a Vicon system. The second is a more direct approach, focusing on training the networks using gait descriptors (like step width, stride length, cadence, max. flexions, max. extensions and rotations at the joints, among other descriptors) directly. In this second approach, 23 most important  descriptors  for distinguishing pathological from normal gait were selected for the analysis. This set of descriptors was obtained according to the recommendations of clinicians Molina and Carratal\'a \cite{FranciscoMolinaRueda:2020}. 

The outline of this paper is organized as follows. Section 2 presents a state-of-the-art review of current methods applied to improve the accuracy of 3D cameras. Section 3 is dedicated to the methodology. This section explains the data collection protocol, as well as the process followed for the definition of joint kinematics. Section 4 addresses the two training approaches and the system accuracy achieved in each. Section 5 is focused on the comparison of both training approaches. Section 6 discusses the results of this study by comparing them with the current approaches. Finally, Section 7 concludes this study. 

\section{State of the art}
\label{sec:state_of_the_art}
The accuracy of 3D cameras in gait analysis has been a highly discussed topic.  For example, Springer and Seligmann \cite{Springer:2016} presented a review of 12 studies that assessed gait analysis with a Microsoft Kinect sensor and a gold standard system. Results of this review indicated good validity for only some spatio-temporal parameters, and not enough validity for gait kinematics variables. For this reason, different approaches have pursued to improve the accuracy of 3D cameras in gait analysis. These approaches have covered, among others, the fusion of 3D cameras with inertial sensors \cite{BersamiraJ:2019}, fusion of multiple 3D cameras\cite{YeungKwok:2013, Geerse:2015,Muller:2017}, or the use of regression equations \cite{AminiAmin:2019} to increase accuracy in the detection of foot-off and foot-contact events. In \cite{Muller:2017}, M\"uller et al. applied a multi-camera configuration that was intended to improve the estimation of spatio-temporal parameters. In the study developed by Matthew et al. \cite{MatthewRobertPeter:2019}, the low accuracy of 3D cameras is attributed to the lack of a gait model for the data retrieved by the sensor, therefore, the authors propose the improvement of the estimations applying a new model based on rigid bodies. Nichols et al. \cite{NicholsJulia:2016} presented another approach based on retro-reflective markers. The researchers found that retro-reflective markers could be used to enhance the motion capture process with 3D cameras.

Another approach has been the fusion with inertial sensors. In the study presented by Alizadegan and Behzadipour \cite{ALIZADEGANALIREZA:2017}, the authors proposed a new method to improve accuracy and real-time performance of inertial joint angle estimation for upper limb rehabilitation applications. The position measurements retrieved from a Kinect sensor were used to correct for the sensor-to-segment misalignment of the inertial sensors. Also in the study presented by Destelle et al. \cite{DestelleFrancois:2014} the authors built an improved skeleton using information fused from a Kinect sensor and nine inertial sensors fixed to the subject's forearms, arms, thighs, shanks and chest. This method proved to obtain more accurate measurements of joint angles, despite a complex calibration process prior to each experiment. 

Previous studies have also attempted sensor latency compensation. Latency compensation allows for the suppression of time delays due to data processing time in 3D cameras. In our previous study \cite{GuffantisensorsDTW:2020}, the latency of the Kinect V2 sensor was measured and corrected using the Dynamic Time Warping (DTW) algorithm. The latency compensation process improves the estimation of joint angles during gait analysis.

ANNs have been another method to improve the accuracy of 3D cameras for gait analysis. For example, in the study of Kidzi\'nski et al. \cite{Kidzin:2019}, ANNs were applied to train the detection of foot-contact and toe-off events. Detection of these events is the initial step in the post-processing of most quantitative gait analysis workflows. Also, in the study presented by Bersamira et al. \cite{BersamiraJ:2019}, the researchers applied ANNs for the integration of 3D cameras and inertial sensors. The authors claim that the fusion of these systems makes it possible to obtain gait data comparable to those produced by a Vicon system. However, as mentioned above, according to Destelle et al. \cite{DestelleFrancois:2014} the use of IMUs together with 3D cameras is an alternative that requires a complex calibration prior to each experiment.

The low accuracy of the measurements obtained using 3D cameras in gait analysis has also been attributed to the joint estimation method. Commonly used Software Development Kits (SDKs) for skeleton tracking estimate the joint center using ML techniques. Firstly, ML is applied to label the pixels corresponding to each body segment. Afterwards, the intersection of these segments is identified to estimate the joint center location \cite{ZhangZheng:2012}. According to Matthew et al. \cite{MatthewRobertPeter:2019}, by applying this method, the estimated joint centers may be biologically inconsistent. This can lead to errors in the location of the ankle, knee, and hip, which complicates their use for later gait analysis \cite{Pfister:2014}. This problem encouraged the development of a series of deep learning methods for joint estimation \cite{ZhouXingyi:2016,HoangVan:2019}. These methods have been particularly applied in self-occlusive tasks such as sitting. Unfortunately, these methods do not study more sophisticated gait descriptors.

In summary, previous studies have attempted to improve the accuracy of the estimation of joint location, but not the gait kinematics itself. The studies that have addressed gait analysis have only dealt with spatio-temporal parameters. However, achieving an accurate enough gait analysis system based on 3D cameras remains a challenge. In contrast to previous studies, this study aims to directly improve the estimation of kinematic gait signals and gait descriptors based on supervised learning through the application of two training approaches. 

\section{Methodology}
\label{sec:methods}

\subsection{Ethics statement}
The current study was approved by the ethics committee of the Faculty of Physical Activity and Sport Sciences INEF (Polytechnic University of Madrid, Spain). The experiments were performed in the Sports Biomechanics Laboratory of INEF. All subjects provided written informed consent prior to participation. In addition, the participants completed an electronic questionnaire with basic questions related to health problems and musculoskeletal injuries suffered in recent years.

\subsection{Data Collection}

Thirty-seven healthy participants (16 female and 21 male, Table~\ref{tab:Demographics}) were involved in the study. A total of 207 gait recordings were processed. 

\begin{table}[htbp]
  \centering
  \caption{Demographics}
  \begin{adjustbox}{max width=0.47\textwidth}
    \begin{tabular}{ll}
    \hline
    \textbf{Participants:} & 37 \\
    \textbf{Gait sequences:} & 207 \\
    \textbf{Male:} & 21 \\
    \textbf{Female:} & 16 \\
    \textbf{Mean age:} & 21$\pm$2 years \\
    \textbf{Mean weight:} & 67.41$\pm$10.28 kg. \\
    \textbf{Mean height:} & 172.99$\pm$8.53 cm. \\
    \hline
    \end{tabular}%
    \end{adjustbox}
  \label{tab:Demographics}%
\end{table}%

\begin{figure}[t]
\centering
\includegraphics[width=0.45\textwidth]{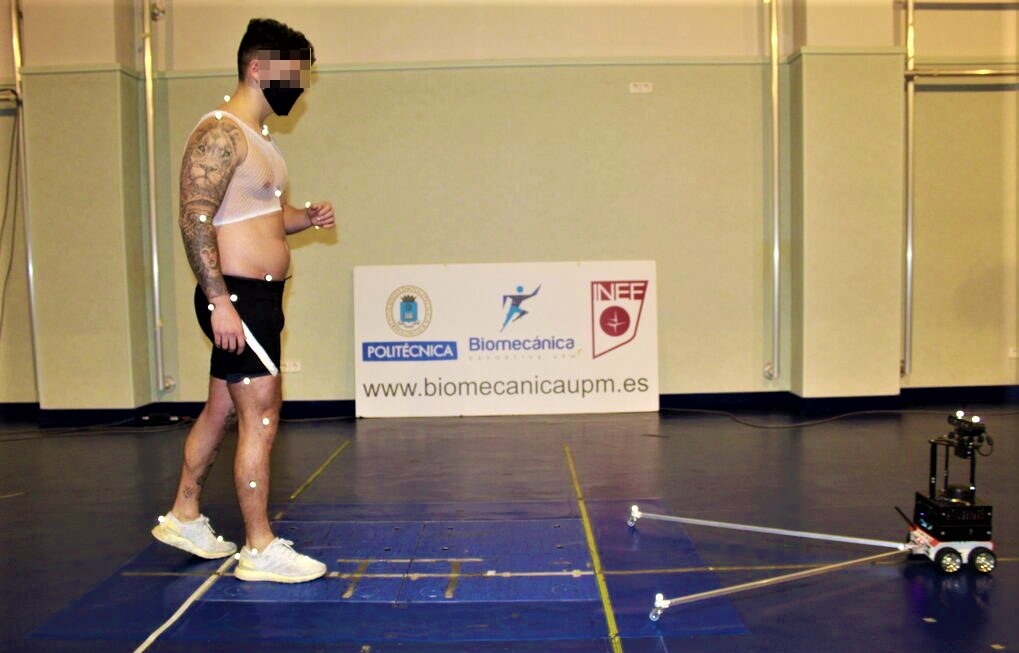}
\caption{An overview of the experimental setup. Faculty of Physical Activity and Sport Sciences INEF (Polytechnic University of Madrid,  Spain).}
\label{fig:experimental_environment}
\end{figure}

The skeleton tracking task was performed with an Orbbec Astra 3D camera using the 0.34.1 release of the Nuitrack SDK. For each gait sequence, 15 joint trajectories were recorded at 30 Hz. During the experiments, the 3D camera was mounted on a mobile robot Rosbot 2.0 from Husarion. The robot is configured to maintain 2.5 meters distance from the person while walking. To achieve this, a controller was implemented in the robot, which was presented in our former study \cite{GUFFANTIroboticsAS:2021}. We  encourage the  reader  to review  the  aforementioned  study  for  more information about the robot configuration and control. The use of the mobile robot allows to obtain a path long enough to analyze the human gait in spite of the limited range of the 3D sensor. Consequently, this avoids the use of multiple sensors and the calibration processes that these configurations require. 
Simultaneously, gait information was retrieved with a Vicon system, which was considered as the ground truth in this study. The Vicon system consisted of six M2 MCAM cameras recording at a sampling rate of 120 Hz. Fig.~\ref{fig:experimental_environment} shows the data collection environment.

The experiments were conducted in two phases, which are summarised below:

\begin{itemize}
    \item Participant preparation (15 minutes). Thirty-two reflective markers were placed on anatomical landmarks following the Plug-in-Gait model \cite{DavisRoy:1991}. The participant wore running shoes, short tights, a top (in the case of women) and socks below the ankles. In addition, the participant signed an informed consent and completed an electronic questionnaire. The height and weight of each participant was registered. Finally, a warm-up and gait speed normalization process was done whereby the subject was prepared to walk at an average speed of 1m/s. 
    \item Gait data capture (15 minutes). Gait acquisition was performed in a 15x5-meter corridor, in a one-way straight line gait. Six iterations per participant were repeated, giving a total of 222 gait sequences. However, some of these were discarded due to the loss of markers in the Vicon system, whose trajectories could not be reconstructed by filling in gaps. As a result, a total of 207 gait recordings were used for the supervised learning stage.
\end{itemize}

\subsection{Data processing}
The robot-mounted 3D camera and the Vicon system started recording data at different times and through different input streams. The timestamp of the first heel strike was used to synchronize the data. This point was retrieved from the anterior-posterior distance between ankle joints (the inter-ankle distance) according to Ceccato et al. \cite{Ceccato:2009}. This event represented the onset of the recording. The end of the recording is determined by the last timestamp registered by the Vicon system. The data from both systems was cut between these limits. Due to the difference in sampling frequencies, the Vicon time series were linearly interpolated using the timestamps of the 3D camera data. This process ensures the same query points. Finally, to minimize any fluctuation, a low pass filter with a cut-off frequency of 4 Hz was applied in the 3D camera data.

\subsection{Data analysis}
To relate as much as possible the gait model applied by the Nuitrack in the 3D camera with that applied by the Plug-in-Gait in the Vicon system, it was necessary to find an equivalence between the landmarks detected by both systems by defining certain mapping rules. This mapping process is not new. Mapping rules similar to those used in \cite{MaMengxuan:2018,TaoWei:2015} were adopted, and are shown in Table~\ref{tab:Mapping_rules}. 

\begin{table}[h]
\caption{Mapping rules to compare the skeleton model of the 3D camera and the Vicon system.}
\begin{adjustbox}{max width=0.47\textwidth}
\centering
  \begin{threeparttable}[t]
  \centering
    \begin{tabular}{ll}
    \hline
    \textbf{3D Camera} & \textbf{Vicon Markers}\\
    \hline
    Spine base & Midpoint [RPSI,LPSI,RASI,LASI]\\
    Spine middle & Midpoint [T10,STRN] \\
    Spine shoulder & Midpoint [C7,CLAV] \\
    Left shoulder & Marker LSHO \\
    Left elbow & Midpoint [LELB marker A,LELB marker B] \\
    Left hand & Midpoint [LWR marker A,LWR marker B] \\
    Right shoulder & Marker RSHO \\
    Right elbow & Midpoint [RELB marker A,RELB marker B] \\
    Right hand & Midpoint [RWR marker A,RWR marker B] \\
    Left hip & Hip joint centering algorithm[LASI,LPSI] \tnote{1} \\
    Left knee & Midpoint [LKNE marker A,LKNE marker B] \\
    Left ankle & Midpoint [LANK marker A,LANK marker B] \\
    Right hip & Hip joint centering algorithm[RASI,RPSI] \tnote{1} \\
    Right knee & Midpoint [RKNE marker A,RKNE marker B] \\
    Right ankle & Midpoint [RANK marker A,RANK marker B]\\
    \hline
    \end{tabular}%
    
    \begin{tablenotes}
     \item[1] The hip joint center location was calculated with the hip joint centering algorithm recommended by \cite{DavisRoy:1991}.
     \end{tablenotes}
     \end{threeparttable}%
     \end{adjustbox}
  \label{tab:Mapping_rules}%
\end{table}%

Six Vicon markers (LTROC, LTHI, LTIB, RTROC, RTHI and RTIB) were not considered in this mapping process. For the analysis of gait in this study, the joint angles from elbow, knee, hip, shoulder, trunk, and pelvis were analyzed in sagittal plane. Furthermore, the joint angles from hip, and shoulder were analyzed in frontal plane. In addition, an examination of trunk and pelvis took place in the transverse plane. 

\subsection{Inverse kinematic model to compute joint angles with the 3D camera}
\label{inverse_kinematics}

\begin{figure*}[htbp]
\centering
\includegraphics[width=1.0\textwidth]{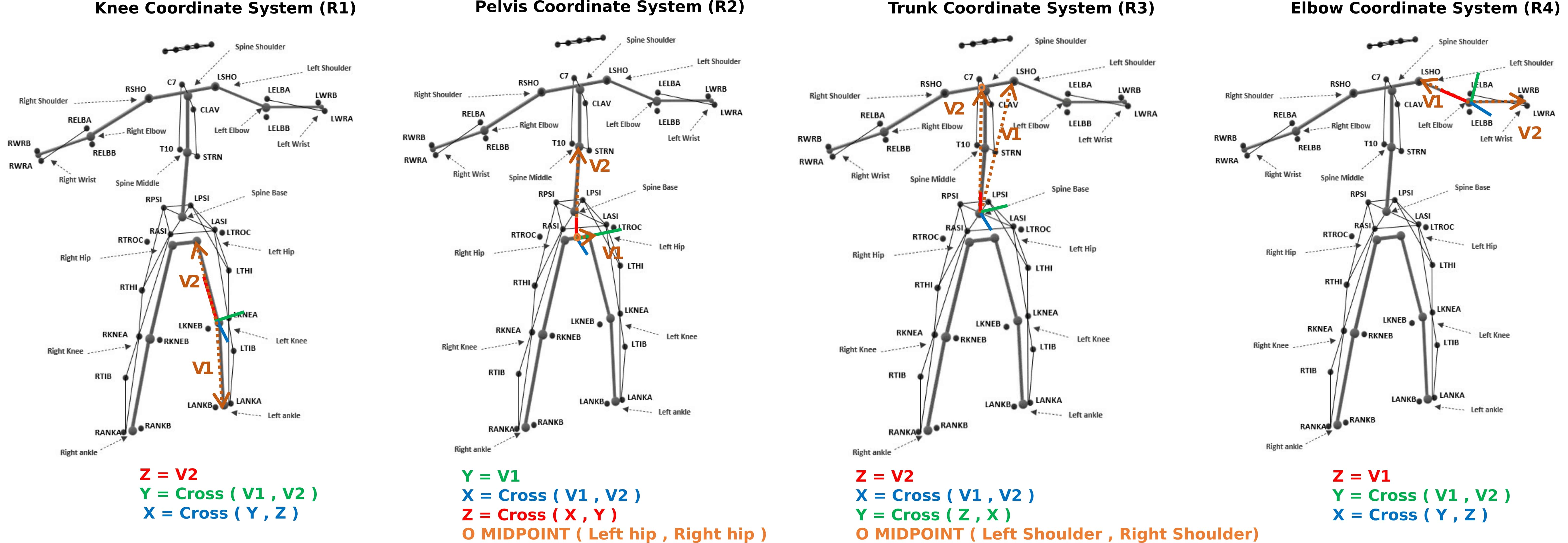}
\caption{Coordinate systems definition of the body segments used to apply the inverse kinematics. Four local coordinate systems are shown. For each joint the coordinate system is indicated where the X-axis is in blue, the Y-axis in green, and the Z-axis in red. }
\label{fig:Coordinate_systems}
\end{figure*}

The inverse kinematics is the process to retrieve joint angles from 3D coordinates of landmarks. For the inverse kinematics process in the 3D camera, the coordinate systems definition of the body segments were compatible with the International Society of Biomechanics recommendations \cite{WuGe:1995}. First, four local coordinate systems are defined at the joints: $R1=$ $^{w}F_{Knee}$, $R2=$ $^{w}F_{Pelvis}$, $R3=$ $^{w}F_{Trunk}$, $R4=$ $^{w}F_{Elbow}$. These are the projected knee, pelvis, trunk, and elbow coordinate frames, on the world frame, and are defined as shown in Fig.~\ref{fig:Coordinate_systems}.

To locate these coordinate systems, vectors V1 and V2 are defined in alignment with the body segments involved. In Fig.~\ref{fig:Coordinate_systems}, the coordinate axes are color-coded: X in blue, Y in green and Z in red. In this figure, at the bottom of each skeleton, the definition of each coordinate axis is indicated according to the order in which they are defined. For example, in the case of the knee coordinate system $R1$, first the Z axis is defined as parallel to the vector V2, which connects knee to hip. Then the Y-axis is defined as the cross product between V1 and V2, where V1 is the vector connecting knee to ankle. Finally, X is defined as the cross product between Y and Z. In some cases, in order to define the coordinate system, it was necessary to calculate a virtual landmark at the midpoint of two adjacent landmarks. These midpoints are colored in orange. For example, the midpoint of left and right hip was used to define the coordinate system $R2$ and the midpoint between left and right shoulder was used to define the coordinate system $R3$.

Once the local coordinate systems are defined, the mathematical definitions of the joint angles correspond as closely as possible to the existing clinical terminology as follows:

\begin{itemize}
  
    \item The hip joint angles were determined as the Cardan rotation angles in the sequence YXZ between the pelvic and knee coordinate systems, i.e. the relative rotation $Eyxz1$, where $Eyxz1=R1\cdot R2^{T}$.
    
    \item The pelvic angles were calculated using the Cardan rotation angles in the sequence YXZ between the world and pelvis coordinate frames. This corresponds to the absolute rotation of $R2$ with respect to the world frame. 
    
    \item The Cardan rotation angles between the world coordinate and the trunk coordinate frames were used to determine the trunk angles. This corresponds to the absolute rotation of $R3$ with respect to the world frame.

    \item The angle between the two vectors crossing at the knee joint center was considered as the knee joint angle. These vectors were rotated according to $R1$.
    
    \item The angle between the two vectors crossing at the elbow joint center was considered as the elbow joint angle. These vectors were rotated according to $R4$.
    
    \item The shoulder joint angles were determined as the Cardan rotation angles in the sequence YXZ between the trunk and elbow coordinate systems, i.e. the relative rotation $Eyxz2$, where $Eyxz2=R4\cdot R3^{T}$.  
    
\end{itemize}

\section{Learning process: Training, Validation and Testing}

The large number of experiments performed together with the robot-mounted 3D camera and the Vicon system allowed the application of supervised learning algorithms to improve the resulting estimations. This application was only possible due to the nature of our study in which the signals can be post-processed before obtaining the final results. There are two basic possible approaches for training and both have been studied in order to see which one achieves a better result: training based on kinematic gait signals, and training based on gait descriptors. 

As mentioned above, 37 participants provided 207 gait recordings. Of these, the recordings corresponding to 20\% of the participants were excluded from the learning process (i.e. were not used for training, validation or testing). This represents the re-testing group of participants. 
The data from these participants was used to make a further comparison between both training approaches. 

\subsection{Training approach based on joint kinematics}

\begin{figure*}[htbp]
\centering
\includegraphics[width=0.95\textwidth]{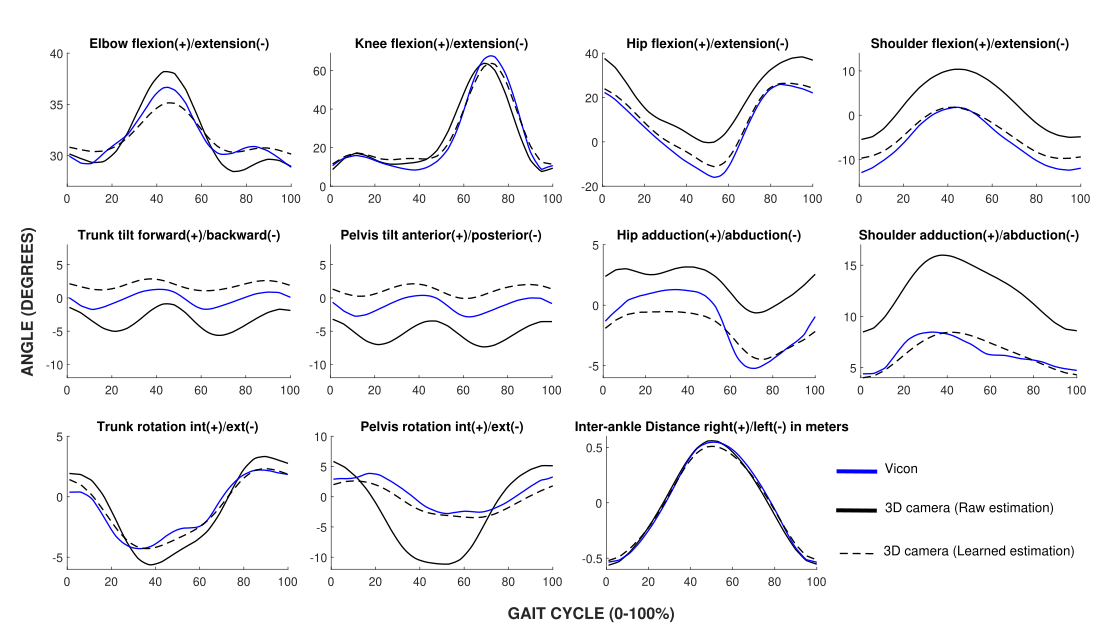}
\caption{Kinematic gait cycles retrieved from the re-testing group of participants. The gait cycles are normalized from 0-100\% and averaged for all the iterations. The figure shows the Vicon system (blue line), the raw estimation of 3D camera (continuous black line), and the learned estimation of 3D camera (dashed black line). }
\label{fig:gait_cycles_V_R_RT}
\end{figure*}

Since the samples from the Vicon system were linearly interpolated using the timestamps of the 3D camera, the samples from both systems are temporally aligned. This allowed us to associate each sample of the camera with a corresponding label on the Vicon signal.

Each kinematic signal was represented by a single vector of dimension Nx1, where N was the total number of samples collected from all the experiments. From this single vector, a training matrix was generated by applying a windowing process. The windowing process is used to split the signal into several frames, or windows. The skip factor between frames was set to one sample. Therefore, the final training matrix had a dimension of (N-W+1)xW, where W was the windowing width. On this basis, each training sample was a vector of 1xW that had a corresponding target of dimension 1x1 coming from the Vicon signal.


This data set was randomly split into three classes: training, validation and testing. The partitioning followed the rule 60/20/20, as follows:
\begin{itemize}
    \item 60\% of the data was used for training.
    \item 20\% of the data was used to validate the network and stop training before overfitting occurs.
    \item 20\% of data was used as a completely independent test of network generalization.
\end{itemize}

Here a two-layer feedforward backpropagation ANN was created and trained per each kinematic signal. The ANN architecture had 10 neurons per layer, an input layer with a number of entries equal to the windowing width factor and an output layer with a single outcome. Experimentally, a windowing width of 3 samples was determined to be the most effective. The training algorithm updated weight and bias values according to Levenberg-Marquardt optimization. Mean squared error (MSE) was used as the performance function. The maximum number of training iterations was set to 1000 epochs.  

Table~\ref{tab:Error_rates_kinematic_signals_test} summarizes the accuracy of the system before and after training. When analyzing the correlations shown in Table~\ref{tab:Error_rates_kinematic_signals_test}, it can be seen that before training, trunk tilt, pelvis tilt, hip add./abd., shoulder add./abd. and pelvis rotation had correlations lower than 0.70. We can attribute the low accuracy in shoulder and hip add./abd to the principle of operation of 3D cameras.  Since the 3D camera does not provide useful information for the detection of motion in frontal plane, the SDK relies on the intrinsic parameters of the RGB camera for the estimation, so these motions are approximations of medium quality. However, after training, the improvements are remarkable. Just  trunk tilt and pelvis tilt remain with correlation lower than 0.70. Despite this fact, considering the low range of motion of these joints (about 2 degrees), the results are satisfactory.
 
\begin{table}[t]
  \centering
  \caption{Accuracy of the estimation provided by the 3D camera in sagittal, frontal and transverse planes before and after training. Results are evaluated using the root-mean-square error (RMSE) and the Pearson Correlation (r). Values are displayed in the format mean$\pm$SD.\\}
  \begin{adjustbox}{max width=0.47\textwidth}
    \begin{tabular}{lcccc}
    \cline{2-5}      
         {}    & \multicolumn{2}{c}{\textbf{BEFORE TRAINING}} &  \multicolumn{2}{c}{\textbf{AFTER TRAINING}}\\
         {}    & {\textbf{RMSE}} &  {\textbf{r}} & {\textbf{RMSE}} &  {\textbf{r}}\\
         
    \hline
     
    Elbow flex/ext.($\degree$) & 3.54 $\pm$ 0.18 & 0.79 $\pm$ 0.02 & 2.03 $\pm$ 0.08 & 0.80 $\pm$ 0.02\\
    Knee flex/ext.($\degree$) & 6.89 $\pm$ 0.29 & 0.94 $\pm$ 0.01 & 5.94 $\pm$ 0.31 & 0.95 $\pm$ 0.01 \\
    Hip flex/ext.($\degree$) & 11.38 $\pm$ 0.24 & 0.93 $\pm$ 0.01 & 4.52 $\pm$ 0.19 & 0.95 $\pm$ 0.00 \\
    Shoulder flex/ext.($\degree$) & 10.03 $\pm$ 0.28 & 0.69 $\pm$ 0.03 & 6.48 $\pm$ 0.40 & 0.71 $\pm$ 0.03 \\
    Trunk tilt($\degree$) & 5.39 $\pm$ 0.11 & 0.45 $\pm$ 0.04 & 1.94 $\pm$ 0.06 & 0.55 $\pm$ 0.03 \\
    Pelvis tilt($\degree$) & 5.81 $\pm$ 0.11 & 0.43 $\pm$ 0.04 & 1.91 $\pm$ 0.06 & 0.55 $\pm$ 0.03 \\
    Hip add/abd.($\degree$) & 4.79 $\pm$ 0.11 & 0.64 $\pm$ 0.03 & 2.59 $\pm$ 0.10 & 0.70 $\pm$ 0.03 \\
    Shoulder add/abd.($\degree$) & 7.65 $\pm$ 0.22 & 0.68 $\pm$ 0.03 & 3.27 $\pm$ 0.25 & 0.73 $\pm$ 0.03 \\
    Trunk rotation($\degree$) & 2.77 $\pm$ 0.15 & 0.82 $\pm$ 0.02 & 1.72 $\pm$ 0.11 & 0.87 $\pm$ 0.02 \\
    Pelvis rotation($\degree$) & 6.95 $\pm$ 0.18 & 0.62 $\pm$ 0.03 & 2.71 $\pm$ 0.09 & 0.71 $\pm$ 0.02 \\
     
    \hline
    \end{tabular}%
  \end{adjustbox}
  \label{tab:Error_rates_kinematic_signals_test}%
\end{table}%

On the other hand, joint kinematics could be affected by the difference in the gait models between Vicon and the 3D camera. Since the number of markers detected by the camera is smaller than those detected by the Vicon, the definition of the coordinate systems of the body segments is in fact different in each system. While the Vicon uses four markers to detect the kinematics of the pelvis, the 3D camera uses only two landmarks, plus one belonging to the middle spine. However, the improvement in the correlation of the signals shown in Table~\ref{tab:Error_rates_kinematic_signals_test}, reflects a fit of the models. Thus,  the kinematic signals reported by the 3D camera more closely resembles that of the Vicon. 

Fig.~\ref{fig:gait_cycles_V_R_RT} shows the effects of the training process for kinematic signals of gait in the re-testing group of participants. From the observation of Fig.~\ref{fig:gait_cycles_V_R_RT}, another problem arising from the difference in gait models can be noted. In some of the signals, such as the hip flex./ext., shoulder flex./ext., trunk tilt, pelvis tilt and shoulder add./abd., there is an offset with respect to the signal of the Vicon system. The correction of this offset is another benefit of the learning process that provides a better interpretation of joint kinematics. In the case of the trunk and pelvis tilt, this offset could not be corrected completely. The randomness of this offset in the training data and the small range of motion of these joints (about 2 degrees) made it difficult for the ANN to correct it. However, a considerable improvement in the correlation of these signals was obtained.

It is very important to emphasize that the training is not aimed at learning just the normal gait pattern performed by the participants. This would make the system unsuitable for the analysis of new gait patterns or pathological patterns. On the contrary, what the ANN actually learns is how to correct the measurement errors of the original gait model in the 3D camera, based on the examples provided by the Vicon system. To achieve this, the ANN does not receive as input the whole gait cycle in each sample. Instead, it receives each the information into data frames thanks to the windowing process. This is extremely important for generalizing learning. Consequently, it is possible to apply the training result to the analysis of new gait patterns.

\subsection{Training approach based on gait descriptors}

\begin{figure}[htb]
\centering
\includegraphics[width=0.47\textwidth]{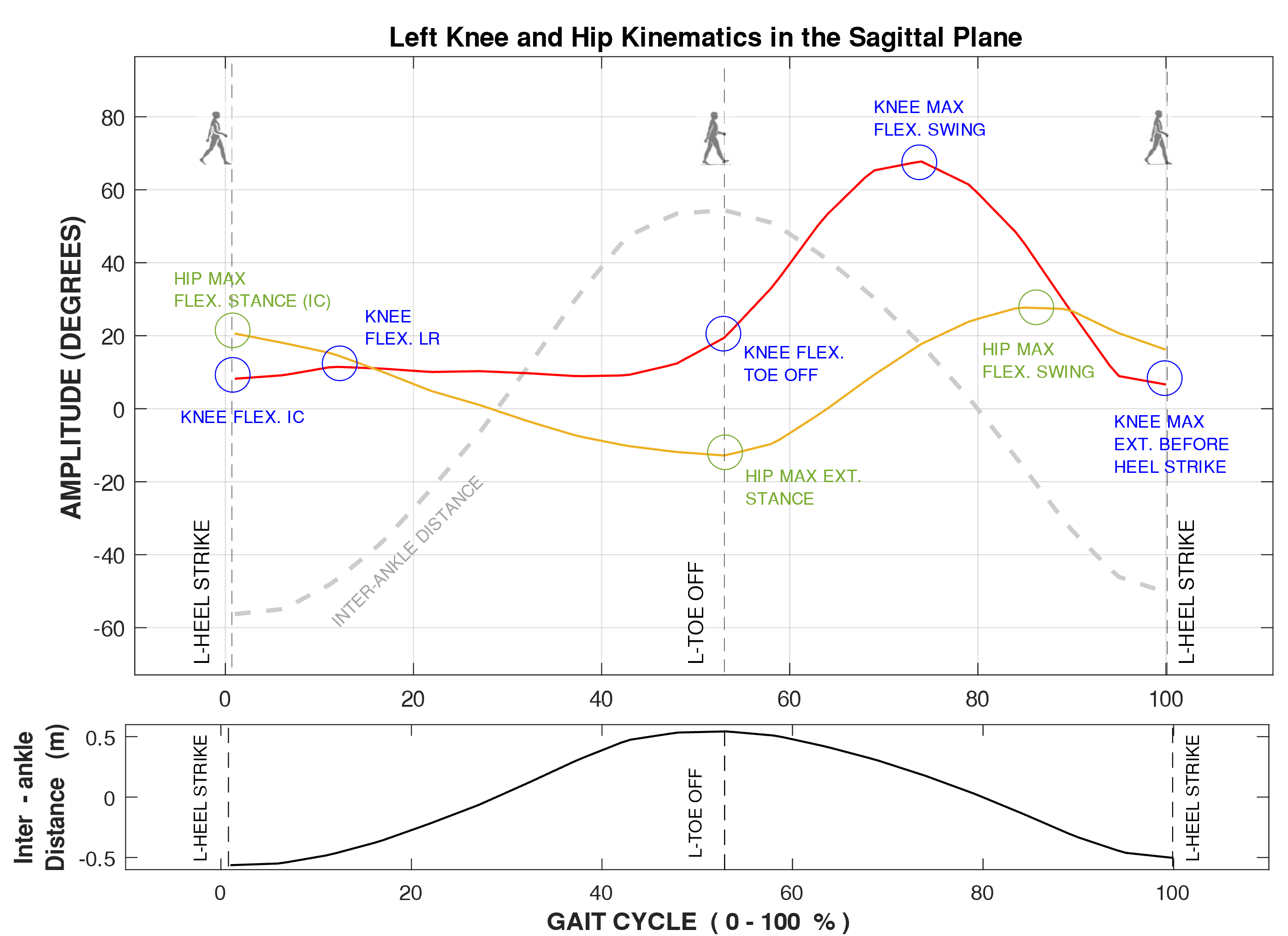}
\caption{Identification of kinematic descriptors for left knee and hip in sagittal plane. The reference signal for the detection of heel strike and toe-off events was the inter-ankle distance \cite{Ceccato:2009}.}
\label{fig:Knee_and_Hip_Kinematics_in_the_Sagittal_Plane_1}
\end{figure}

This second approach arises from the concern of whether it would be better to train the ANN directly with gait descriptors rather than with kinematic signals divided into data frames. This is a more direct way to teach the ANN to report gait descriptors more accurately. In order to apply this second approach, following an in-depth study of gait disturbances in neurological patients, and on the recommendation of clinicians Molina and Carratal\'a \cite{FranciscoMolinaRueda:2020}, 27 most important descriptors representing the main differentiators of normal and pathological gait were selected for the analysis. As listed in Table~\ref{tab:Error_rates_gait_events_test}, 11 were spatio-temporal descriptors and 16 were kinematic descriptors. 

Fig.~\ref{fig:Knee_and_Hip_Kinematics_in_the_Sagittal_Plane_1} illustrates the process followed for the identification of the set of kinematic gait descriptors. According to Ceccato et al. \cite{Ceccato:2009}, the reference signal for the detection of heel strike and toe-off events was the maximal anterior-posterior distance between ankle joints (or inter-ankle distance). In this signal there is a characteristic sinusoidal curve when the X coordinate of the foot markers is graphed versus time. In this curve, the valleys correlate to the time at which the foot comes into contact with the ground (the heel strike) and the peaks fit with the initiation of swing phase (the toe-off) \cite{ZeniJA:2007}. Fig.~\ref{fig:Knee_and_Hip_Kinematics_in_the_Sagittal_Plane_1} illustrates the process followed for the identification of kinematic descriptors of knee and hip in sagittal plane. The rest of the kinematic descriptors correspond to instants of maxima and minima in the corresponding kinematic signals. In addition, the set of spatio-temporal descriptors were calculated according to established definitions \cite{Winter:2009,BeauchetOlivier:2009,Muller:2017}.

\begin{table}[t]
  \centering
  \caption{Accuracy of the system for the detection of 11 spatio-temporal and 16 kinematic descriptors. The table shows the results before and after training. Results are evaluated using the root-mean-square error (RMSE). Values are displayed in the format mean$\pm$SD.\\}
  \begin{adjustbox}{max width=0.47\textwidth}
    \begin{tabular}{lcc}
    \hline
      \multirow{2}{*}{\textbf{GAIT DESCRIPTORS}} & {\textbf{BEFORE TRAINING}} & {\textbf{AFTER TRAINING }}\\
      \textbf{} & \textbf{RMSE} &  \textbf{RMSE}\\
    \hline
    \textbf{SPATIO-TEMPORAL DESCRIPTORS}\\
    step width ($m$) & 0.022 $\pm$ 0.01 & 0.021 $\pm$ 0.01 \\ 
    left stride length ($m$) & 0.15 $\pm$ 0.02 & 0.07 $\pm$ 0.01 \\ 
    right stride length ($m$) & 0.13 $\pm$ 0.03 & 0.05 $\pm$ 0.01 \\
    left stride time ($s$) & 0.03 $\pm$ 0.01 & 0.04 $\pm$ 0.01 \\
    right stride time ($s$) & 0.06 $\pm$ 0.02 & 0.06 $\pm$ 0.02 \\
    right step time ($s$) & 0.04 $\pm$ 0.02 & 0.03 $\pm$ 0.01 \\
    left step time ($s$) & 0.027 $\pm$ 0.01 & 0.026 $\pm$ 0.01 \\
    right cadence ($steps/min$) & 4.65 $\pm$ 1.08 & 4.60 $\pm$ 1.27 \\
    left cadence ($steps/min$) & 3.60 $\pm$ 1.80 & 4.01 $\pm$ 1.51 \\
    percentage of foot stance ($\%$) & 3.74 $\pm$ 0.78 & 2.74 $\pm$ 0.73 \\
    percentage of foot swing ($\%$) & 2.73 $\pm$ 1.41 & 1.32 $\pm$ 0.29 \\
    \textbf{KINEMATIC DESCRIPTORS}\\
    trunk max. tilt ($\degree$) & 4.22 $\pm$ 0.87 & 1.68 $\pm$ 0.45 \\
    trunk min. tilt ($\degree$) & 5.93 $\pm$ 1.21 & 2.18 $\pm$ 0.35 \\
    pelvis max. tilt ($\degree$) & 4.31 $\pm$ 0.57 & 1.33 $\pm$ 0.52 \\
  
    pelvis min. tilt ($\degree$) & 6.14 $\pm$ 0.64 & 1.67 $\pm$ 0.31 \\
    hip max. adduction ($\degree$) & 3.86 $\pm$ 0.75 & 1.37 $\pm$ 0.35 \\
    hip min. abduction ($\degree$) & 5.02 $\pm$ 0.54 & 2.22 $\pm$ 0.67 \\
    pelvis max. rotation ($\degree$) & 6.72 $\pm$ 1.59 & 2.96 $\pm$ 0.97 \\
    pelvis min. rotation ($\degree$) & 9.81 $\pm$ 0.98 & 2.02 $\pm$ 0.47 \\
    hip max. extension during stance ($\degree$) & 14.08 $\pm$ 1.53 & 3.60 $\pm$ 0.83 \\
    hip max. flexion during swing ($\degree$) & 12.22 $\pm$ 1.31 & 2.61 $\pm$ 0.54 \\
    hip max. flexion during stance ($\degree$) & 15.10 $\pm$ 1.47 & 3.70 $\pm$ 1.05 \\
    knee initial contact position ($\degree$) & 7.43 $\pm$ 1.97 & 4.91 $\pm$ 1.33 \\
    knee position at toe-off ($\degree$) & 9.85 $\pm$ 1.81 & 3.29 $\pm$ 1.03 \\
    knee max. flexion in load response ($\degree$) & 4.10 $\pm$ 1.02 & 4.52 $\pm$ 1.03 \\
    knee max. flexion during swing ($\degree$) & 4.06 $\pm$ 1.21 & 3.39 $\pm$ 1.69 \\
    knee max. extension before heel strike ($\degree$) & 5.76 $\pm$ 1.68 & 4.81 $\pm$ 1.52\\
    \hline
    \end{tabular}%
    \end{adjustbox}
  \label{tab:Error_rates_gait_events_test}%
\end{table}%

Each gait descriptor was represented by a single vector of dimension Px1, where P was the number of gait cycles collected from all the experiments. This means that one sample of the set of descriptors is extracted for each gait cycle. Given that 27 descriptors were analyzed, a training matrix of dimension Px27 was generated. A target matrix of the same dimensions was obtained with the estimations of the Vicon system. Corresponding columns were used to train independent neural networks. Each ANN used a feedforward backpropagation architecture, with two hidden layers, 10 neurons per layer, an input layer with a single entry and an output layer with a single outcome. The training algorithm updated weight and bias values according to Levenberg-Marquardt optimization, and MSE was used as the performance function. The maximum number of training iterations was set to 1000 epochs. Again the data set was split into three classes: training, validation and testing, following the rule 60/20/20 (60\% for training, 20\% for validation and 20\% for testing).

Table~\ref{tab:Error_rates_gait_events_test} summarizes the results of this training approach. Results show that the spatio-temporal parameters have no further improvement after the training process. An exception occurs in the case of left/right stride length, where a noticeable correction is observed. Regarding the sixteen kinematic descriptors, the results are better after training. Especially the descriptors for hip, pelvis and trunk show remarkable improvements.

\section{Comparison of training approaches}
\label{sec:results}

\begin{figure}[htb]
\centering
\includegraphics[width=0.45\textwidth]{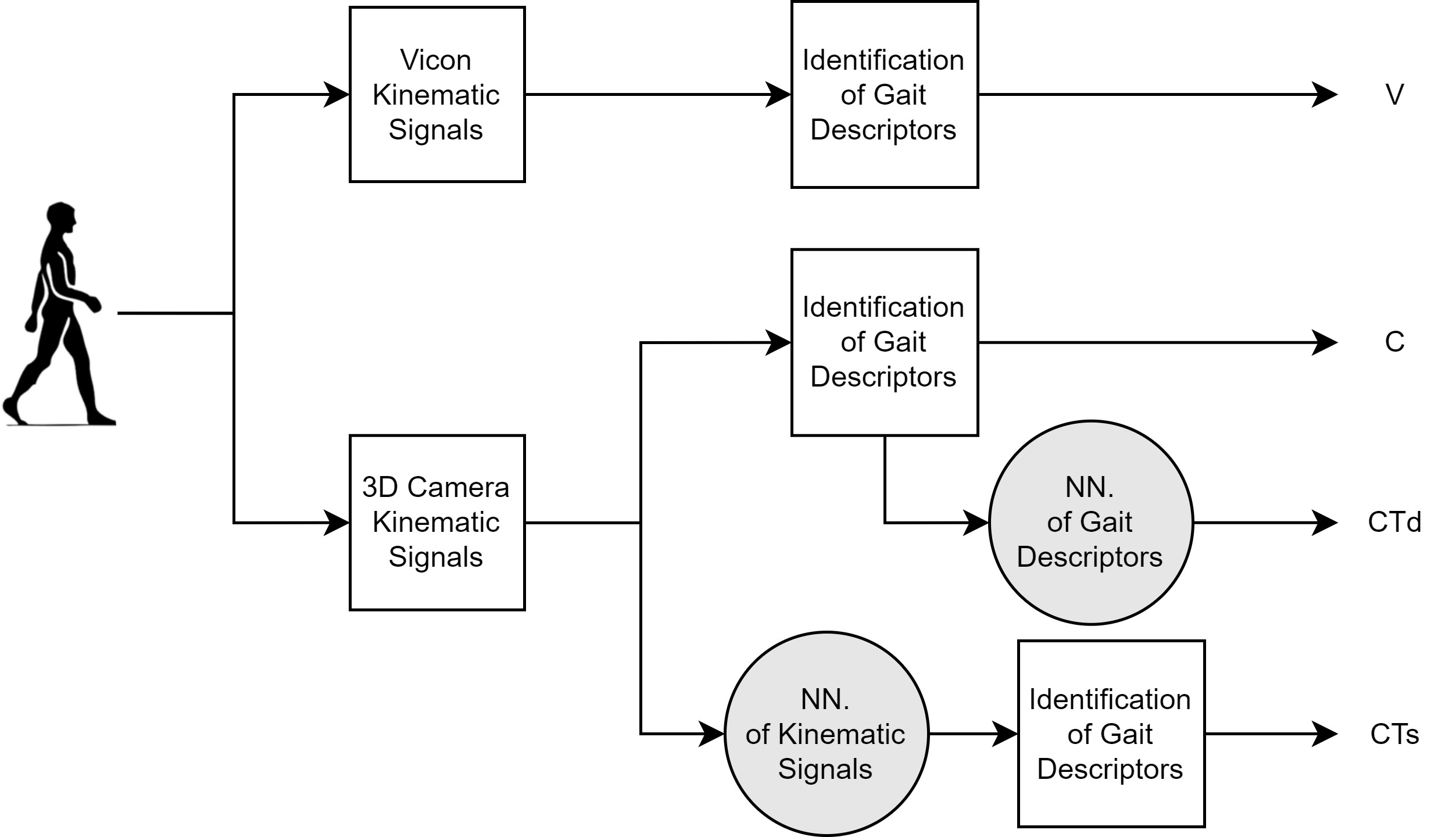}
\caption{Process followed to obtain the data for the comparison of the two training approaches. }
\label{fig:Comparison_of_trainings}
\end{figure}

\begin{figure*}[htbp]
\centering
\includegraphics[width=0.95\textwidth]{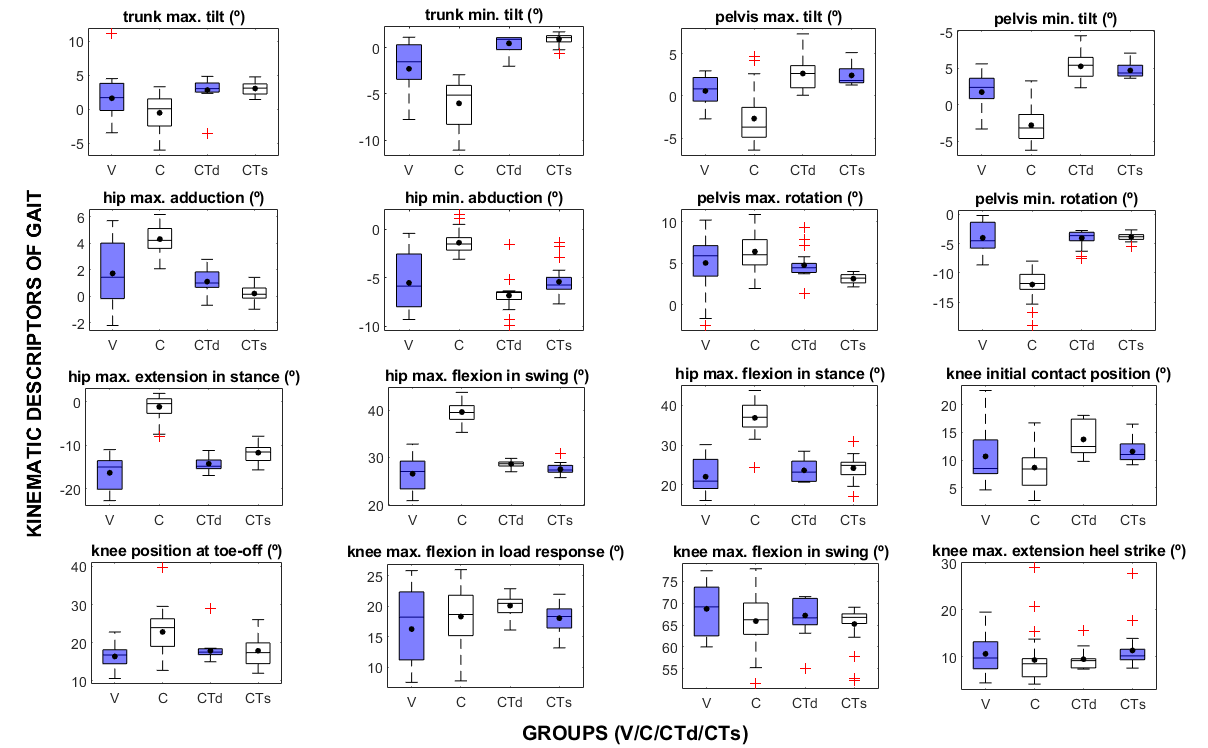}
\caption{Results of the comparison of both training approaches. The box plots show the 25th percentile, 75th percentile, median, minimum and maximum values. The mean value is represented with a black dot. The box plot closest to the Vicon is colored in blue, which is also colored in blue. }
\label{fig:kinematic_events_boxplot_RTevents_RTsignals}
\end{figure*}

This section is focused on the comparison of both training approaches. Specifically, it compares the accuracy for the estimation of the sixteen kinematic descriptors of gait. For this purpose, the re-test group of participants was used. Fig.~\ref{fig:Comparison_of_trainings} shows the process followed to obtain the data for comparison. 
On the basis of this process, four estimations can be compared: the Vicon estimation (V); the 3D camera raw estimation (C); the 3D camera trained with kinematic signals (CTs); and the 3D camera trained with gait descriptors (CTd).

Results of this comparison are shown in the 
Fig.~\ref{fig:kinematic_events_boxplot_RTevents_RTsignals}. A visual representation of the data is shown using box plots. These box plots summarize the 25th percentile, 75th percentile, median, minimum and maximum values of each group of data. The mean value is also shown with a black dot over the box plots. In this figure, the box plot belonging to the system with mean value closest to that of the Vicon is colored in blue, which is also colored in blue.



When analyzing the sixteen kinematic descriptors shown in Fig.~\ref{fig:kinematic_events_boxplot_RTevents_RTsignals}, it is clear that the trained systems (CTd or CTs) are always better than the untrained system (C). When comparing the two training approaches, it is observed that in 62\% of the events analyzed (10/16) the approach based on gait descriptors (CTd) is better than the approach based on kinematic signals (CTs), where only in 38\% of the cases (6/16) it proved to be better. However, it can be observed that for events of large amplitude, such as maximum knee and hip flexions and extensions, training based on gait descriptors (CTd) is better. On the contrary, for kinematics of smaller range of motion, the training based on kinematic signals (CTs) is superior. This can be noted in the pelvis maximum and minimum tilt, and in the lower amplitude descriptors of knee kinematics such as the knee initial contact position, knee maximum flexion in load response, and in the knee maximum extension before heel strike. Therefore, it is not possible to define which training approach is completely better. Despite the fact that training based on gait descriptors (CTd) presented higher accuracy in most parameters, training based on kinematic signals (CTs) has an enormous advantage because it allows to know the complete kinematics of the joint, which is very useful when determining deviations between gait patterns for clinical purposes. Consequently, the selection of the training approach will depend on the purpose of the study conducted.

\section{Discussion}
\label{sec:discussion}

It is interesting to compare the results obtained with other techniques for improving the estimations of 3D cameras. An in-depth comparison can be made by analyzing Table~\ref{tab:Comparison_of_accuracy_improvement}. Current approaches have attempted to improve the accuracy of 3D cameras in gait analysis through remodeling the skeleton, sensor fusion, or applying machine learning techniques. 
In the study presented by M\"uller et al. \cite{Muller:2017}, a validation of enhanced Kinect sensor for gait assessment was presented. The idea of that study was to compute a spatially averaged skeleton tracked by two Kinect sensors placed at both sides of the walking path. The study demonstrated that sensor setups tracking the person only from one-side is less accurate and should be replaced by two-sided setups. 
To fuse data from both sensors, the authors used the iterative closest point method, which adds imprecision to the model. Similarly, the use of more than one camera suggests a requirement of a dedicated room where the system can be setup and left undisturbed between sessions.  

Regarding the fusion of 3D cameras with inertial sensors (IMUs), it is interesting to mention the study presented by Destelle et al. \cite{DestelleFrancois:2014}. The authors built a fused Kinect/IMUs skeleton where the Kinect sensor provided the positions of the joints and the IMUs provided the orientation information. The use of IMUs together with 3D cameras required a complex calibration prior to each experiment. The authors reported a final accuracy of 9.00$\degree$ for elbow flex./ext., and 7.89$\degree$ for knee flex./ext. In contrast, in our study considerably higher accuracies were achieved after training (2.03$\degree$ for elbow flex./ext. and 5.94$\degree$ for knee flex./ext.). It should also be noted that the study of Destelle et al. \cite{DestelleFrancois:2014} was performed for a simulated kicking motion while in our study a real gait was analyzed.

\begin{table*}[htbp]
  \centering
  \caption{Comparison of our study with current approaches.}
  \begin{adjustbox}{width=1.0\textwidth,totalheight=7cm}
  \begin{threeparttable}[t]
  
    \begin{tabular}{lcccccccccccccc}
    
\cline{2-15} & \multicolumn{2}{c}{\textbf{OUR SYSTEM\tnote{1}}} & \multicolumn{2}{c}{\textbf{RIGID BODIES\tnote{2}}} & \multicolumn{2}{c}{\textbf{DTW\tnote{3}}} & \multicolumn{2}{c}{\textbf{MULTI-SENSOR\tnote{4}}} & \multicolumn{2}{c}{\textbf{MULTI-SENSOR\tnote{5}}} & \multicolumn{2}{c}{\textbf{3D CAMERA+IMU\tnote{6}}} & \multicolumn{2}{c}{\textbf{3D CAMERA+IMU\tnote{7}}} \\
\cline{2-15}    & \multicolumn{14}{c}{\textbf{ROOT MEAN SQUARE ERROR}} \\ 
      & \multicolumn{1}{c}{\textbf{BEF.}} & \textbf{AFT.} & \multicolumn{1}{c}{\textbf{BEF.}} & \textbf{AFT.} & \multicolumn{1}{c}{\textbf{BEF.}} & \textbf{AFT.} & \multicolumn{1}{c}{\textbf{BEF.}} & \textbf{AFT.} & \multicolumn{1}{c}{\textbf{BEF.}} & \textbf{AFT.} & \multicolumn{1}{c}{\textbf{BEF.}} & \textbf{AFT.} & \multicolumn{1}{c}{\textbf{BEF.}} & \textbf{AFT.} \\
\cline{1-15} 
      \textbf{Elbow flex./ext.} & 3.54 & 2.03 & - & - & - & - & - & - & - & - & - & 5.90 & 13.33 & 9.00 \\
      \textbf{Knee flex./ext.} & 6.89 & 5.94 & 9.40 & 5.20 & 13.15 & 10.47 & - & - & - & - & - & - & 28.23 & 7.89 \\
      \textbf{Hip flex./ext} & 11.38 & 4.52 & 11.10 & 5.60 & 6.20 & 2.65 & - & - & - & - & - & - & - & - \\
      \textbf{Shoulder flex./ext.} & 10.03 & 6.48 & - & - & 9.53 & 5.80 & - & - & - & - & - & 4.70 & - & - \\
      \textbf{Trunk forw./back. tilt} & 5.39 & 1.94 & 3.90 & 3.50 & 2.09 & 1.93 & - & - & - & - & - & - & - & - \\
      \textbf{Pelvis ant./post. tilt} & 5.81 & 1.91 & - & 6.10 & 3.22 & 3.14 & - & - & - & - & - & - & - & - \\
      \textbf{Hip add./abd.} & 4.79 & 2.59 & - & - & 5.83 & 4.70 & - & - & - & - & - & - & - & - \\
      \textbf{Shoulder add./abd.} & 7.65 & 3.27 & - & - & 11.76 & 11.69 & - & - & - & - & - & - & - & - \\
      \textbf{Trunk int./ext. rot.} & 2.77 & 1.72 & - & - & 5.26 & 5.09 & - & - & - & - & - & - & - & - \\
      \textbf{Pelvis int./ext. rot.} & 6.95 & 2.71 & - & - & 5.88 & 4.81 & - & - & - & - & - & - & - & - \\
    \hline
      \textbf{step width (m)} & 0.022 & 0.021 & - & - & - & - & - & 0.01 & 0.0066 & 0.0039 & - & - & - & - \\
      \textbf{AV. stride length (m)} & 0.14 & 0.06 & - & - & - & - & - & 0.00 & 0.0004 & 0.0020 & - & - & - & - \\
      \textbf{AV. stride time (s)} & 0.05 & 0.05 & - & - & - & - & - & 0.01 & - & - & - & - & - & - \\
      \textbf{AV.step time (s)} & 0.034 & 0.028 & - & - & - & - & - & 0.01 & 0.000 & 0.000 & - & - & - & - \\
      \textbf{AV. Cadence (steps/min)} & 4.12 & 4.31 & - & - & - & - & - & 0.90 & - & - &   &   &   &  \\
    \hline
    \end{tabular}%
    \begin{tablenotes}
    \item[1] Our trained robot-mounted 3D camera.
    \item[2] Matthew et al.\cite{MatthewRobertPeter:2019}. Single Kinect, tested in a sit to stand task. The original joint data is transformed into a new model built with rigid bodies.
    \item[3] Guffanti et al.\cite{GuffantisensorsDTW:2020}. Correction of latency usind DTW(Dynamic Time Warping).
    \item[4] Geerse et al.\cite{Geerse:2015}. Multi-Kinect v2, 10-Meter Walkway for spatio-temporal gait analysis.
    \item[5] M\"uller et al. \cite{Muller:2017}. Multi-Kinect v2, 7-Meter Walkway for spatio-temporal gait analysis.
    \item[6] Alizadegan and Behzadipour\cite{ALIZADEGANALIREZA:2017}. Single Kinect to improve accuracy of inertial joint angle estimation in upper limb stroke rehabilitation.
    \item[7] Destelle et al.\cite{DestelleFrancois:2014}. Single Kinect in front of the person and fusion with IMUs. The RMSE values were averaged for the two experiments performed by the authors. 
    \end{tablenotes}
    \end{threeparttable}%
    \end{adjustbox}
  \label{tab:Comparison_of_accuracy_improvement}%
\end{table*}%

In the study presented by Alizadegan and Behzadipour \cite{ALIZADEGANALIREZA:2017}, the authors proposed a new method to improve accuracy of IMUs for joint angle estimation in upper limb rehabilitation applications. To correct for the sensor-to-segment misalignment of the inertial sensors, position measurements retrieved from Kinect sensor were used. The experiment was performed on a mechanical upper limb stroke rehabilitation device also known as shoulder wheel. The shaft rotation angle was encoded and this measurement was used as ground truth. The authors reported accuracies of 5.90$\degree$ for elbow flex./ext. and 4.70$\degree$ for shoulder flex./ext. Conversely, in our study, the elbow is reported with higher accuracy (3.54$\degree$ for elbow flex./ext.) but the shoulder is slightly less accurate (6.48$\degree$ for shoulder flex./ext.). It should also be noted that the experimental basis in the study of Alizadegan and Behzadipour \cite{ALIZADEGANALIREZA:2017} is different from ours. While in that study the person remains static, in our study the person is in constant walking motion.

Matthew et al. \cite{MatthewRobertPeter:2019} proposed the improvement of the estimations provided by 3D cameras by remodeling the skeleton to a new one based on rigid bodies. The idea of the authors was to rescale the skeleton detected by the 3D camera, using the height of the person. Then, the joint centers are recalculated based on anthropometric approximations. The method was tested only for a sit to stand task (STS), and the results obtained were satisfactory. However, it should be noted that, applying joint center relocation based on anthropometric approximations may not be a robust technique. This is because 3D cameras are affected by many other factors like light conditions, sensor position, occlusion, among others \cite{LemkensW:2013}. This makes it difficult to relocate joint centers using only anthropometric approximations. Thus, this method may not be effective in all experimental conditions. In addition, our method based on ML obtained better results mainly for trunk and pelvis tilt. While in the study of Matthew et al. \cite{MatthewRobertPeter:2019} errors of 3.50$\degree$ for trunk tilt and 6.10$\degree$ for pelvis tilt were obtained, in our study 1.94$\degree$ and 1.91$\degree$ were achieved respectively.

\setlength{\parskip}{0cm}
\setlength{\parindent}{1em}
    
Finally, in our previous study \cite{GuffantisensorsDTW:2020}, latency correction was proposed using Dynamic Time Warping (DTW). Latency of 3D cameras produces non-symmetrical temporal variations in signals that directly affects the system performance. Results showed a noticeable improvement in system performance. However, these corrections only solve the problem of the temporal disturbances produced by the variable latency of the sensor. Instead, the application of machine learning not only corrects these distortions. This can be observed in the improvement of the correlation in Table~\ref{tab:Error_rates_kinematic_signals_test} and in the correction of waveforms in Fig.~\ref{fig:gait_cycles_V_R_RT}.
In addition to the above, machine learning can also correct discrepancies in the gait model applied by the 3D camera with respect to the Plug-in-gait model applied by the Vicon system. This effect can be observed in Fig.~\ref{fig:gait_cycles_V_R_RT}, where the offsets with respect to the Vicon signals were corrected.

\section{Conclusion}
\label{sec:conclusion}

In this study, it was possible to improve the accuracy of 3D cameras for human gait analysis applications. This was achieved by post-processing raw estimations of the 3D camera using ANNs trained with the data provided by a certified Vicon system. The neural networks were trained based on two different criteria: training based on kinematic gait signals, and training based on gait descriptors. In the latter approach, 11 spatio-temporal and 16 kinematic descriptors representing the main differentiators of normal and pathological gait were selected for the analysis. 

The accuracy of the estimations with the 3D camera was measured before and after training. Results showed lower errors and higher correlations with respect to the gold standard, thus, improving the obtained biomechanical model with the 3D camera. The accuracy obtained when detecting the main descriptors of a pathological gait also showed a substantial improvement, mainly for kinematic descriptors. 


When comparing both training approaches, it was not possible to define which was the absolute best. Regardless,  training  based  on  kinematic  signals  has an enormous  advantage because it allows to know the complete kinematics of the joint. This is very useful when determining deviations between gait patterns along a long-term clinical analysis. Therefore, we believe that the selection of the training approach will depend on the purpose of the study to be conducted.

This study reveals the great potential of 3D cameras and encourages the research community to continue exploring their use in gait analysis.

\section*{Conflicts of interest}
None of the authors have any conflicts of interest associated with this research.

\section*{Acknowledgments}
The research leading to these results has received funding from PID2020-118299RB-I00 - SISTEMA ROBOTICO NO INVASIVO PARA EL ANALISIS BIOMECANICO DE LA MARCHA HUMANA -, ``Convocatoria 2020 Proyectos de I+D+i - RTI Tipo B'', of the Government of Spain.
The authors would like to thank Faculty of Physical Activity and Sports Sciences - INEF, UPM, for the use of Sports Biomechanics Laboratory. 

\bibliography{LIBRARY_GAIT_ANALYSIS}

\end{document}